\documentclass[11pt]{article}
\usepackage{amsfonts}
\usepackage{array}
\usepackage{bbm}
\usepackage{bigstrut}
\usepackage{blindtext}
\usepackage{booktabs}
\usepackage{cleveref}
\usepackage{coling2018}
\usepackage{color}
\usepackage{cprotect}
\usepackage{enumitem}
\usepackage{float}
\usepackage{graphicx}
\usepackage{hyperref}
\usepackage{latexsym}
\usepackage{mathrsfs}
\usepackage{microtype}
\usepackage{multirow}
\usepackage{nccmath}
\usepackage{nicefrac}
\usepackage{siunitx}
\usepackage[justification=centering]{subcaption}
\usepackage{tabularx}
\usepackage{tikz-network}
\usepackage{times}
\usepackage{url}
\usepackage{verbatim}
\usepackage{wrapfig, lipsum}
\usepackage[labelfont=bf, format=plain, justification=justified, singlelinecheck=false]{caption}
\newcolumntype{M}[1]{>{\centering\arraybackslash}m{#1}}
\usepackage{placeins}
\usepackage[utf8]{inputenc} % allow utf-8 input
\usepackage[T1]{fontenc}    % use 8-bit T1 fonts

\title{Investigating the Working of Text Classifiers}

\author{Devendra Singh Sachan$^{\spadesuit,\triangle}$, Manzil Zaheer$^{\spadesuit}$, Ruslan Salakhutdinov$^{\spadesuit}$ \\
$^{\spadesuit}$School of Computer Science, Carnegie Mellon University\\
$^{\triangle}$Petuum, Inc\\
Pittsburgh, PA, USA \\
{\tt \{dsachan, manzilz, rsalakhu\}@andrew.cmu.edu}\\
}

\date{}
\begin{document}
\maketitle
\begin{abstract}
Text classification is one of the most widely studied tasks in natural language processing. Motivated by the principle of compositionality, large multilayer neural network models have been employed for this task in an attempt to effectively utilize the constituent expressions.
%use the word order in sentences.  
Almost all of the reported work train large networks using discriminative approaches, which come with a caveat of no proper capacity control, as they tend to latch on to any signal that may not generalize. Using various recent state-of-the-art approaches for text classification, we explore whether these models actually learn to compose the meaning of the sentences or still just focus on some keywords or lexicons for classifying the document. To test our hypothesis, we carefully construct datasets where the training and test splits have no direct overlap of such lexicons, but overall language structure would be similar. We study various text classifiers and observe that there is a big performance drop on these datasets. Finally, we show that even simple models with our proposed regularization techniques, which disincentivize focusing on key lexicons, can substantially improve classification accuracy.
\end{abstract}

\section{Introduction}

\blfootnote{
    % % final paper: en-us version 
    %
    \hspace{-0.65cm}  % space normally used by the marker
    This work is licensed under a Creative Commons 
    Attribution 4.0 International License.
    License details:
    \url{http://creativecommons.org/licenses/by/4.0/}
}

Text classification is one of the fundamental tasks in natural language processing (NLP) in which the objective is to categorize text documents into one of the predefined classes. This task has a lot of applications such as topic classification of news articles, sentiment analysis of reviews, email filtering, etc. Text classification still remains a significant challenge in language understanding as we need to encode the intrinsic grammatical relations between sentences in the semantic meaning of a document. This is especially crucial for sentiment classification because relations like ``contrast'' and ``cause'' can have great influence on determining the meaning and the overall polarity of a document.

Traditionally, for text classification, bag-of-words model~\cite{harris1954distributional} was used to represent a document. This simple approach uses the term frequency as features followed by a classifier such as na\"ive Bayes~\cite{mccallum1998naive} or support vector machine (SVM)~\cite{joachims1998text}. One drawback of this approach is that it ignores the word order and grammatical structure. It also suffers from data sparsity problem when the training set's size is small but has shown to give good results when size is not an issue~\cite{wang2012baselines}. Before applying these methods, feature selection is typically carried out to reduce the effective vocabulary size by removing the noisy features~\cite{Manning:2008:IIR:1394399}. A key property of these linear classifiers is that they assign high weights to some class label specific keywords, which are also known as lexicons.

The next generation of approaches includes neural networks that have shown to outperform bag-of-words models in text classification tasks~\cite{kim2014convolutional,johnson2016supervised}. These methods typically use multiple layers of convolutional neural network (CNN)~\cite{lecun1998gradient} and/or long short-term memory (LSTM) networks~\cite{hochreiter1997long}. The motivation of using these complex neural network approaches for classification tasks comes from the principle of compositionality~\cite{frege1948sense}, which states that the meaning of a longer expression (e.g. a sentence or a document) depends on the meaning of its constituents. It is believed that lower layers of the network learn representations of words or phrases, and as we move up the layers more complex expressions are represented~\cite{peters2018deep}.

However, we believe that these state-of-the-art text classification techniques do not actually follow this mechanism, despite the motivations. Like any discriminative approach which can pick up any signal, it appears that these techniques most likely still learn and rely heavily on key lexical items or phrases and just use these lexicons to classify the document, which might not generalize to new documents.

To test our hypothesis, we first construct datasets (Section~\ref{sec:data}) where the training and test splits have no direct overlap of such key lexicons while taking care of class imbalance, although remaining language structure remains the same. Such dataset split also occurs in many real-world classification problems. For example, in the case of scientific documents, with the discovery of a new phenomenon/law/theorem, a new keyword is born. The document containing the new word would still belong to one of the existing classes, such as optics/biochemistry/discrete math, and a text classifier should be able to correctly classify the given document as the language structure would remain the same. 
%pretty standard of that field.
%%%
% Not relevant here, diverts from main point, hampers readability 
%%%
%due to limited availability of training data, we may want to use distance supervision to create noisy labeled data for training text classifiers. 
%For all the categories of interest, we can extract all associated lexicons and keywords from knowledge bases and use them to annotate unlabeled data. Since there may be many sentences containing a given lexicon, we can potentially use very large numbers of noisy features that are combined in the resulting classifier. But when used for inference, such classifiers may be required to predict on a test instance whose keywords were unseen during model training.

We study the performance of various text classifiers on this new training-test split and compare performance results with the commonly used random training-test split of such a dataset (Section~\ref{sec:exp}). We observe that there is a big performance gap in current approaches between the two splits. 
%on both such splits of a dataset.
We further show that even simple regularization techniques of replacing key lexicons with random embeddings can improve performance on the training-test split where there is no overlap of such keywords. We also present a novel approach based on the gradient of word embeddings that leads to further improvement on such dataset split (Section~\ref{sec:method}). We also provide two large-scale text classification datasets which contain both the random split and lexicon based split.

We report additional experimental results by modeling the above problem as that of unsupervised domain adaptation where the data distribution of training domain and test domain are different~\cite{ganin2016domain,zhang2017aspect}. Even such domain adaptation approaches do not yield better results than our proposed regularization techniques (Section~\ref{sec:exp}).

\section{Background and Related Work}\label{sec:related}
In this section, we first discuss work related to neural network approaches for text classification followed by a discussion on recent domain adaptation techniques which can be thought of as an alternative solution for our problem.

\subsection{Neural Network Approaches}
Recent research on text classification tasks makes use of neural networks which has shown to outperform methods based on the bag-of-words model. These approaches take distributed representation of words as input which is also known as word vectors~\cite{mikolov2013distributed}. These word vectors can be learned either using skip-gram~\cite{mikolov2013distributed} or Glove~\cite{pennington2014glove} methods. One remarkable property of these vectors is that they learn the semantic relationships between words such that in the embedding space, semantically similar words are closer together. To learn document embeddings from these word vectors,~\newcite{le2014distributed} use distributed bag-of-words (DBOW) approach also known as paragraph vectors while~\newcite{joulin2017bag} computes the average of the word and subword vectors of a document to train a linear classifier.

To extract sentence level features for text classification task,~\newcite{kim2014convolutional} uses shallow CNN with max-pooling on top of pre-trained word vectors. They also observe that learning task-specific vectors through fine-tuning leads to better classification accuracy. Similarly, LSTM network pre-trained using language model parameters or sequence autoencoder parameters is used by~\newcite{dai2015semi} for various text classification tasks. Recently, it was shown by~\newcite{Johnson2015EffectiveUO} that a CNN with dynamic max pooling layer can effectively use the word order structure when trained using one-hot encoding representation of input words. They also learn multi-view region embeddings through semi-supervised learning and incorporate them as additional features to further improve the model's performance~\cite{johnsonNIPS2015}. Similarly, they also perform semi-supervised experiments using a simplified LSTM model which also takes one-hot encoding of words as input~\cite{johnson2016supervised}. 
% Furthermore, adversarial training methods \cite{Miyato ICLR 2017} can be adopted to increase their robustness against noise performance of the text classifiers.

\begin{figure}[t]
\centering
\begin{minipage}{.5\linewidth}
  \centering
  \includegraphics[scale=1.0]{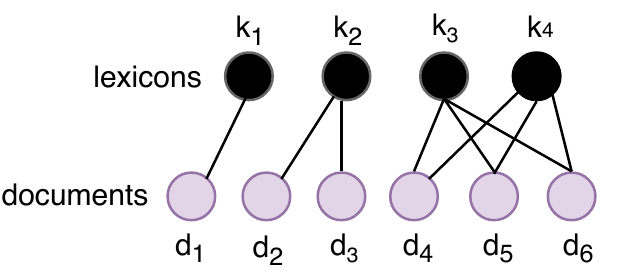}
  \subcaption{}
  \label{fig:model_stepa}
\end{minipage}%
\begin{minipage}{.5\linewidth}
  \centering
  \includegraphics[scale=1.0]{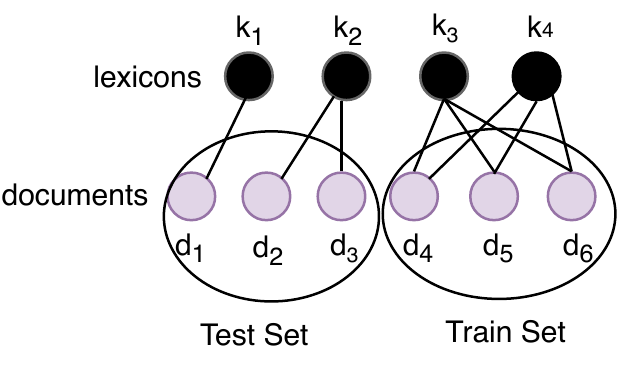}
  \subcaption{}
  \label{fig:model_stepb}
\end{minipage}
\caption{Schematic diagram illustrating the methodology of lexicon dataset construction. In (a), we construct a bipartite, undirected graph between the identified keywords ($k_1, \dots, k_4$) and all the documents ($d_1, \dots, d_6$). We identify all the connected components in this graph. In (b), all the documents belonging to the largest connected component ($d_4, d_5, d_6$) are selected as the training set while those documents belonging to all the other connected components ($d_1, d_2, d_3$) are selected as the test set.}
\label{fig:model}
\end{figure}

% \begin{figure}[t]
% \centering
% \hfill
% \begin{subfigure}{.45\textwidth}
%   \includegraphics[width=\textwidth,trim=0 0 0 0,clip]{images/graph_1.pdf}
%   \captionsetup{justification=centering}
%   \caption{}
%   \label{fig:model_stepa}
% \end{subfigure}
% \hfill
% \begin{subfigure}{0.45\textwidth}
%   \includegraphics[width=\textwidth,trim=0 0 0 0,clip]{images/graph_2.pdf}
%   \captionsetup{justification=centering}
%   \caption{}
%   \label{fig:model_stepb}
% \end{subfigure}
% \hfill
% \caption{Schematic diagram illustrating the methodology of lexicon dataset construction. In (a), we construct a bipartite, undirected graph between the identified keywords ($k_1, \dots, k_4$) and all the documents ($d_1, \dots, d_6$). We identify all the connected components in this graph. In (b), all the documents belonging to the largest connected component ($d_4, d_5, d_6$) are selected as the training set while those documents belonging to all the other connected components ($d_1, d_2, d_3$) are selected as the test set.}
% \label{fig:model}
% \end{figure}

\subsection{Domain Adaptation}
Next, we discuss the recent works applicable to domain adaptation of text classifiers. In unsupervised domain adaptation, the labeled instances from training data are considered as the source domain while unlabeled instances from test data are considered as the target domain. In this paper, it is assumed that there exists covariate shift between the two domains such that the conditional distribution of class label remains the same in both the domains while the marginal distribution of instances may differ~\cite{shimodaira2000improving,bickel2007discriminative}. To address the problem of covariate shift, attempts have been made by adversarially training the encoder so as to make document representation domain invariant~\cite{ganin2016domain}.

Another approach to domain adaptation is multi-task learning as it can help improve the performance of text classifiers on lexicon dataset~\cite{Caruana:1997:ML:262868.262872}. It is assumed that by learning related tasks in parallel while using a shared representation of document encoder can improve generalization by inducing proper inductive biases.

\section{Lexicon Dataset Construction}\label{sec:data}
As mentioned previously, to test our hypothesis	whether the text classifiers just focus on key lexicons, we consider three text corpora: IMDB reviews, its standard subset \cite{maas-EtAl:2011:ACL-HLT2011}, and Arxiv abstracts. For each dataset, we construct two versions: random split version and lexicon split version. In the random split version, selection of training and test examples is done by random sampling. The ratio of test to training examples is kept approximately the same for both the random and lexicon version of each dataset. 
In the lexicon split version, we make sure that key lexicons in training and test examples do not overlap while taking care of class imbalance. Below, we provide stepwise details of the approach used for the construction of lexicon version of the datasets.

\begin{enumerate}
\item \textbf{Identification of important label specific lexicons:} We begin by extracting \emph{tf-idf} weighted, unigram to five-gram word features for all the documents. Using thus obtained feature vectors as input, we train a multinomial na\"ive Bayes classifier for predicting the class on the entire dataset (both training and test examples).
From the trained na\"ive Bayes classifier, we extract the feature weights. 
Next, we rescale the feature weights by dividing all the feature values by the corresponding minimum value for that feature across all the classes. Lastly, to identify lexicons, we select the top-\emph{k} features with the maximum weight for every class. We set the value of \emph{k} to be 1500 for IMDB datasets and 150 for Arxiv abstracts dataset.
We also experiment with Logistic Regression and SVM classifiers but observe that multinomial na\"ive Bayes selects the most diverse set of features. 

\item \textbf{Creation of lexicons-documents graph:} Since each label specific lexicon term can occur in multiple documents, and one document can contain many such lexicons, in order to create a disjoint partition of lexicons and documents into training/test sets, it is natural to represent documents and lexicons using a graph-based structure. Therefore, for each class, we create an undirected graph whose nodes are lexicons and documents respectively. If a lexicon occurs in a document, we create an edge in the graph between the two corresponding nodes. The resulting graph is bipartite, as edges only exist between the lexicon nodes and the document nodes but not among each other (Figure~\ref{fig:model_stepa}).

\item \textbf{Graph partitioning to generate training/test splits:} We compute all the connected components of the lexicons-documents graph. We then identify the largest connected component, i.e. one which contains the maximum number of nodes and select all the documents in it as the training set.\footnote{We also tried spectral partitioning of lexicons-documents graph but it didn't generate balanced graph partitions.} The remaining documents from all the other connected components are selected as the test set (Figure~\ref{fig:model_stepb}). If the ratio of the number of test to training documents is below a cutoff threshold, we repeat the above steps by gradually selecting fewer top-\emph{k} features (i.e. smaller set of lexicons). Some example lexicons for training/test splits of Arxiv abstracts and IMDB reviews datasets is shown in Table~\ref{table:lexicons}.
\end{enumerate}

We want to emphasize that after the creation of training and test sets, the majority of the words appear in both the sets for the lexicon version.

\begin{table}[t]
\centering
\begin{subtable}[c]{\textwidth}
\small
\begin{tabular}{c | c | c | c | c}
\toprule
& \textbf{cs.AI} & \textbf{cs.IR} & \textbf{cs.CV} & \textbf{cs.RO} \\ [0.5ex]
\midrule
\multirow{4}{*}{\textbf{Train}} & reinforcement learning & information retrieval & computer vision & motion planning \\
& logic programming & recommender systems & image segmentation & humanoid robot \\
& markov decision processes & collaborative filtering & deep convolutional neural & dynamic environments \\
& probabilistic inference & recommendation systems & super resolution & aerial vehicle \\
\midrule
\multirow{3}{*}{\textbf{Test}} & bayesian networks & matrix factorization & pose estimation & multi robot \\
& graphical models & social networks & optical flow & simultaneous localization \\
& constraint satisfaction & search engines & sparse representation & extended kalman filter \\
\bottomrule
\end{tabular}
\captionsetup{justification=centering}
\caption{Arxiv abstracts}
\end{subtable}

\begin{subtable}[c]{\textwidth}
\medskip
\small
\begin{tabular}{c | c | c | c | c | c}
\toprule
& \textbf{most-positive} & \textbf{positive} & \textbf{neutral} & \textbf{negative} & \textbf{most-negative}\\ [0.5ex]
\midrule
\multirow{3}{1.7em}{\textbf{Train}} & absolutely amazing & completely satisfied & average movie & disappointed & a huge let down \\
& awesome experience & good experience & is just ok & scope for improvement & movie was awful \\
& fantastic movie & great movie & moderate movie & movie sucks & very frustrating \\
\midrule
\multirow{2}{1.5em}{\textbf{Test}} & outstanding performance & nice experience & satisfactory movie & not a good & worst experience \\
& very impressive & movie is good & avg performance & no plot in movie & pretty bad \\
\bottomrule
\end{tabular}
\captionsetup{justification=centering}
\caption{IMDB reviews}
\end{subtable}

\caption{Example lexicons for training and test set of Arxiv abstracts and IMDB reviews datasets. The column headers in both the tables indicate the names of various classes in these datasets. For Arxiv abstracts dataset, the details of all the class names can be found in the URL: \url{https://arxiv.org}}
\label{table:lexicons}
\end{table}

\section{Methods}\label{sec:method}
In this section, we will first briefly describe a simple neural network for text classification on which our proposed regularization methods are based. %and then motivate our proposed methods.

Let the vocabulary size be $V$ and embedding dimension be $D$. Our network consists of an embedding layer ($E \in \mathbb{R}^{V \times D}$), single layer bidirectional LSTM (BiLSTM)~\cite{schuster1997bidirectional}, a pooling layer, and finally a linear layer with softmax function for classification. The sequence of word ids ($w_t, t \in [1, T]$) are given as input to the embedding layer which maps them to dense vectors. The forward LSTM and backward LSTM of a BiLSTM processes these input sequence of word vectors in forward and reverse directions respectively. The hidden state of forward LSTM and backward LSTM is concatenated at every time-step and are passed to the pooling layer which computes the maximum value over time dimension to obtain the representation of the input sequence~\cite{Conneau2017SupervisedLO}. We train the model by minimizing the cross-entropy loss.

While training various neural network models on both random and lexicon splits of ACL IMDB dataset~\cite{maas-EtAl:2011:ACL-HLT2011}, we observed that on the lexicon split, the network is easily able to learn the training data distribution in 1-2 epochs. These models also overfit on the lexicon split as the accuracy gap between the training and test set widens up. We hypothesize that this happens because, during training step, the model is able to memorize common label-specific keywords occurring in the training set. During evaluation, when the model is not able to spot such keywords in the test set, as it contains non-overlapping partition of keywords from the training set, the performance degrades quite rapidly as compared to the random split version. Therefore, motivated by these observations, we propose two methods which attempt to prevent neural networks from memorizing the word order structure in keywords by introducing more randomness in them.

\subsection{Keyword Anonymization}
In order to prevent the model from memorizing keyword specific rules and thus learning degenerate representations, we introduce more randomness in our training data split~\cite{DBLP:conf/nips/HermannKGEKSB15}. In the first step, we identify some keywords with high scores using a supervised classifier trained using bag-of-words as features. The first step is similar to the first step of data construction, with the important distinction that now we only operate on training set as test set is assumed to be unknown during the training phase. In the second step, we anonymize the corpus by replacing a single word selected at random from the occurrences of such keyword phrases with a placeholder word `\verb+ANON+'. 
% An example of the second step is shown in Table~\ref{table:anonymization}. 
In the third step, we assign a random word embedding during model training for every occurrence of the placeholder word `\verb+ANON+'. These modified word embeddings are given as input to the neural network encoder.

In this way, we corrupt the information present in keywords by introducing some form of random noise in the data. This is one of the ways to regularize model training. We later show that this regularization forces the network to learn context-aware text representations which gives a significant gain in classification accuracy. Motivated by the effectiveness of this method, we now propose an end-to-end version as the next approach.

\subsection{Adaptive Word Dropout}
This approach is based on embedding layer's gradient which is computed at every optimization step. The motivation of this approach is based on the observation that key lexicon terms have a higher norm of the embedding gradient. 
In this method, we first compute the word embedding gradient ($\nabla E \in \mathbb{R}^{V \times D}$) during the backward step. In the second step, for every word we maintain the running average ($A(w)$) of the L2-norm of this gradient. This average is computed with respect to the number of completed optimization steps.
\begin{align*}
\tilde A(w) &= \tilde A(w) + \sqrt{\sum_{j=1}^D \nabla E_{ij}^2} \\
A(w) &= \frac{1}{\text{optim\_steps}} \tilde A(w)
\end{align*}
In the third step, we compute the dropout probability of every word as:
\begin{align*}
\begin{aligned}
P_d(w) &= 1 - \sqrt{\frac{t}{A(w)}}
\end{aligned}
\end{align*}
In the above equation, $t$ is a hyperparameter (typical values of $t$ are $10^{-3}$, $10^{-4}$). We apply dropout on the words before the word embedding layer with probability $P_d(w)$~\cite{DBLP:journals/jmlr/SrivastavaHKSS14}. We also apply variational dropout~\cite{NIPS2016_6241} on the output of word embedding layer.

\section{Experiments}\label{sec:exp}
We now present empirical studies in order to establish that (i) many text classifiers resort to just identifying key lexicons and thus poorly performing on specially crafted dataset splits (Section~\ref{sec:res}), and (ii) simple regularization techniques which disincentivize focusing on key lexicons can significantly boost performance (Section~\ref{sec:res_reg}).

\subsection{Dataset Description}
To illustrate aforementioned claims, we evaluate on an existing dataset for binary sentiment classification and also collected two more large-scale text classification datasets. The details of these three datasets are described below:

\begin{itemize}
\item \textbf{ACL IMDB}: This is a popular benchmark dataset~\cite{maas-EtAl:2011:ACL-HLT2011} of IMDB movie reviews for coarse-grained sentiment analysis in which the task is to determine if a review belongs to the positive class or to the negative class. The original random split version of this dataset contains an equal number of highly positive and highly negative reviews. To construct its lexicon based version, we apply our approach to the combined training and test splits of this dataset.

\item \textbf{IMDB reviews}: This is a much bigger version of the IMDB movie reviews dataset in which the task is to do fine-grained sentiment analysis. We collect more than 2.5 million reviews from IMDB website and partition them into five classes based on their ratings out of 10. These classes are \emph{most-negative}, \emph{negative}, \emph{neutral}, \emph{positive}, and \emph{most-positive}.

\item \textbf{Arxiv abstracts}: In this, the task is to do multiclass topic classification. We construct this dataset by collecting more than 1 million abstracts of scientific papers from ``\emph{arxiv.org}''. Each paper has one primary category such as cs.AI, stat.ML, etc. which we use as its class label. We selected those primary categories which have at least 500 papers. In order to extract text data, we use the title and abstract of each paper. In both the Arxiv abstracts and IMDB reviews dataset, we keep the ratio of the number of instances in the test set to that of training set as 0.6.
\end{itemize}
We pre-process the data by converting all the text to lowercase followed by Penn Treebank style tokenization. An overall summary of the training/test sets of lexicon and randomized splits is provided in Table~\ref{table:statistics}.

% TABLE: DATASETS
\begin{table}[t]
\centering
%\medskip
  \begin{tabular}{@{} M{0.3cm}  r r | r r | r r | r r | r r | r r  >{\bfseries}S[table-format=3.2,table-number-alignment=right]@{}}
  \toprule
    & \multicolumn{4}{c|} {\textbf{Arxiv abstracts}} & \multicolumn{4}{c|} {\textbf{IMDB reviews}} & \multicolumn{4}{c} {\textbf{ACL IMDB}} \\
    \cmidrule{2-13}
     & \multicolumn{2}{c|} {lexicons} & \multicolumn{2}{c|} {random} & \multicolumn{2}{c|} {lexicons} & \multicolumn{2}{c|} {random} & \multicolumn{2}{c|} {lexicons} & \multicolumn{2}{c} {random} \\
    \cmidrule{2-13}
    & Train & Test & Train & Test & Train & Test & Train & Test & Train & Test & Train & Test \\
\midrule
    $c$ & 127 & 127 & 127 & 127 & 5 & 5 & 5 & 5 & 2 & 2 & 2 & 2\\
    $l$ & 162 & 140 & 153 & 153 & 299 & 252 & 278 & 281 & 290 & 254 & 234 & 228 \\
    $N$ & 668K & 438K & 664K & 443K & 1.48M & 1.07M & 1.49M & 1.07M & 23K & 26K & 25K & 25K \\
    $V$ & 297K & 238K & 301K & 244K & 590K & 509K & 601K & 505K & 75K & 78K & 75K & 74K \\
\bottomrule
\end{tabular}
\caption{This table shows the dataset summary statistics. $c$: Number of classes, $l$: Average length of a sentence, $N$: Dataset size, $V$: Vocabulary size}
\label{table:statistics}
\end{table}

\subsection{Experimental Setup}
In this section, we will discuss our experimental setup. For simple baseline comparisons, we conduct experiments with methods which take bag-of-words as input feature representations. Specifically, we compute \emph{tf-idf} weighted \emph{n-gram} features for each document to train multinomial na\"ive Bayes and Logistic Regression classifiers. For neural network models, we have a common experimental setup based on Pytorch framework~\cite{paszke2017automatic} for all the datasets unless specified otherwise. Due to computational constraints, we did not perform extensive hyperparameter tuning for the methods considered.

We select most common 80K words for model training. We initialize the embedding layer parameters for all the models using 300-dimensional pre-trained embeddings. These embeddings are trained using skip-gram approach~\cite{mikolov2013distributed} on the combined training and test set.\footnote{We use \emph{word2vec} tool from \url{https://code.google.com/p/word2vec}} The embeddings of words which are not present in these vectors are uniformly initialized. The hidden state of LSTM and BiLSTM has 1024 and 512 dimensions respectively. For training of CNN models, we follow the settings mentioned in~\newcite{kim2014convolutional}. We perform model training using mini-batch stochastic gradient descent with a batch size of 150. Optimization is performed using Adam Optimizer~\cite{DBLP:journals/corr/KingmaB14} with default parameter settings. To train LSTM models, we perform backpropagation for 250 timesteps on IMDB datasets and 150 timesteps on Arxiv abstracts dataset. For model regularization, we apply dropout~\cite{DBLP:journals/jmlr/SrivastavaHKSS14} with probability 0.5 to the input and output of LSTM. In order to prevent gradient explosion problem, we perform gradient clipping~\cite{DBLP:conf/icml/PascanuMB13} by constraining the norm of the gradient to be less than 5.

% TABLE: RESULTS
\begin{table}[!t]
  \medskip
  \centering
  \begin{tabular}{@{} l | c c c | c c c | c c c >{\bfseries}S[table-format=3.2,table-number-alignment=right]@{}}
  \toprule
  \textbf{Model} & \multicolumn{3}{c|}{\bf{Arxiv abstracts}} & \multicolumn{3}{c|}{\bf{IMDB reviews}} & \multicolumn{3}{c}{\bf{ACL IMDB}}\\
  \midrule
   & random & lexicon  & $\Delta$ & random & lexicon & $\Delta$ & random & lexicon & $\Delta$ \\ 
  \cmidrule{2-10}
  Na\"ive Bayes & 57.6 & 40.4 & 17.2 & 54.9 & 41.6 & 13.3 & 89.8 & 79.3 & 10.5\\
  Logistic Regression & 65.2 & 40.6 & 24.6 & 60.0 & 41.9 & 18.1 & 90.5 & 80.6 & 9.9\\
  \midrule
  DocVec & 62.6 & 38.9 & 23.7 & 55.9 & 44.4 & 11.5 & 88.6 & 82.8 & 5.8 \\
  FastText  & 66.8 & 43.8 & 23.0 & 57.9 & 45.1 & 12.8 & 88.5 & 80.5 & 8.0 \\
  Deep Sets & 55.7 & 37.9 & 17.8 & 52.0 & 46.8 & 5.2 & 88.2 & 79.6 & 8.6\\
  \midrule
  LSTM & 65.0 & 45.3 & 19.7 & 64.6 & 48.6 & 16.0 & 90.6 & 81.9 & 8.7\\
  BiLSTM & \bf{67.8} & 46.5 & 21.3 & \bf{64.8} & 48.9 & 18.9 & \bf{91.4} & 83.1 & 8.3 \\
  CNN-MaxPool & 65.8 & 46.0 & 19.8 & 50.5 & 44.1 & 6.4 & 89.6 & 80.5 & 9.1 \\
  CNN-DynMaxPool & 66.6 & 45.5 & 21.1 & 52.1 & 41.5 & 10.6 & 90.0 & 80.9 & 9.1\\
  \midrule
  Adv-Training & -- & 45.1 & -- & -- & 47.3  & -- & -- & 82.2 & -- \\
  Multi-task Learning & -- & 43.5 & -- & -- & 44.8 & -- & -- & 78.7 & -- \\
  \midrule
  LSTM-Anon & 67.2 & 48.2 & 19.0 & 62.6 & 50.3 & 12.3 & 89.0 & 83.1 & 5.9 \\
  BiLSTM-Anon &  67.5 & 48.7 & 18.8 & 62.8 & 51.4 & 11.4 & 89.7 & 84.1 & 5.6 \\
  \midrule
  Adaptive Dropout & 67.2 & \bf{49.1} & 18.1 & 64.0 & \bf{52.6} & 11.4 & 91.2 & \bf{86.0} & 5.2 \\
  \bottomrule
  \end{tabular}
\caption{Classification accuracy of various models on random and lexicon based version of each dataset. We also show the accuracy difference ($\Delta$) between these two results.
\textbf{Na\"ive Bayes}: \emph{n-gram} feature extraction using \emph{tf-idf} weighting followed by na\"ive Bayes classifier.
\textbf{Logistic Regression}: \emph{n-gram} feature extraction using \emph{tf-idf} weighting followed by Logistic Regression classifier.
\textbf{DocVec}: Document vectors trained using DBOW model \protect\cite{le2014distributed}.
\textbf{FastText}: Average of the word and subword embeddings \protect\cite{joulin2017bag}.
\textbf{Deep Sets} \protect\cite{zaheer2017deep}: two layer MLP on top of word embeddings with max pooling layer. 
\textbf{LSTM}: Word level LSTM as a document encoder in which hidden state of the last time-step is used for classification. 
\textbf{BiLSTM}: Word level bidirectional LSTM as a document encoder in which forward LSTM and backward LSTM hidden states are concatenated followed by max pooling layer \protect\cite{Conneau2017SupervisedLO}. 
\textbf{CNN-MaxPool}: CNN with max pooling layer \protect\cite{kim2014convolutional}. 
\textbf{\mbox{CNN-DynMaxPool}}: CNN with dynamic max pooling layer. For details on dynamic max pooling, we request the reader to refer to \protect\newcite{Johnson2015EffectiveUO}.
\textbf{Adv-Training}: Adversarial training of encoder is done to fool the discriminator and make the representation of training and test instances domain invariant.
\textbf{Multi-task Learning}: A shared BiLSTM encoder is used to perform joint training of text classifier, denoising autoencoder, and adversarial training.
\textbf{LSTM-Anon}: LSTM encoder applied to the anonymized training data. \textbf{BiLSTM-Anon}: BiLSTM encoder applied to the anonymized training data.
\textbf{Adaptive Dropout}: BiLSTM encoder applied after embedding gradient-based adaptive word dropout.}
\label{table:results}
\end{table}

\subsection{Results}\label{sec:res}
In order to estimate the difficulty level on both the lexicon and random splits version of a dataset, we do experiments with a wide range of popular approaches and show their results in Table~\ref{table:results}. It can be observed from the results that for each dataset, there is a big performance gap between the random split and lexicon split for all the models. In this section, we will analyze these performance results. During our analysis, we will mostly focus on classifier's performance on the lexicon version of a dataset. For a detailed analysis of the performance of a method on the random version, we encourage the reader to read the associated paper.

\subsubsection{Bag-of-Words Model}\label{sec:res_bow}
Here, we analyze the performance of methods which takes bag-of-words representation as input. We observe that the simple generative classifier of na\"ive Bayes performs poorly as compared to other approaches on both Arxiv abstracts and IMDB reviews dataset, while its results are competitive on ACL IMDB dataset. Because of its relatively low scores, we don't experiment with more complex generative models. In contrast, a simple discriminative classifier such as Logisitic Regression performs significantly better than na\"ive Bayes on the random version for all the datasets. However, on lexicon version, it's performance is quite low when compared to neural network methods. Due to this, the accuracy gap for this method is largest between both the dataset versions. These low scores of above classifiers on lexicon version can be explained owing to the conditional independence assumption among the input features. During training step, bag-of-words based models assign high weights to class-specific keywords. When this trained model is not able to spot such discriminative keywords in the test set due to strict no overlap condition, the performance of these methods degrades. %Also, the better performance of LR model on random version can be due to the similar distribution of the keywords in both the training and test splits.

\subsubsection{Simple Word Embedding based Methods}\label{sec:res_emb}
Next, we analyze the results of classifiers which are based on simple linear or nonlinear transformation of word embeddings. We train 300-dimensional document vectors~\cite{le2014distributed} using standard training settings for DBOW model.\footnote{We use the existing implementation from the open-source gensim toolkit.} Similarly, we use the recommended training setting for FastText method \cite{joulin2017bag}. These methods further improve the lexicon results in both the IMDB datasets. Although DBOW and FastText methods don't strictly make use of word order in a sentence, improved performance suggests that they leverage the semantic properties of the embedding space as the document vectors are also learned in the same geometrical space as the word vectors.

\subsubsection{Convolutional and Recurrent Networks}\label{sec:res_rcnn}
As discussed previously (Section ~\ref{sec:related}), it has been shown that both CNN and LSTM models can learn effective text representations. We observe that these networks show much better performance on both the random and lexicon version as compared to the above-discussed methods. We also note that the use of a more complex model such as BiLSTM gives performance improvement in both random and lexicon splits for all the datasets. 

Although there is an improvement in classification accuracy, still there is a large drop in performance from random to lexicon split. For the best performing approach of BiLSTM model, accuracy difference between random and lexicon split of ACL IMDB, IMDB reviews, and Arxiv abstracts dataset is 8.3\%, 18.9\%, and 21.3\% respectively. In order to narrow down this performance gap, there is a need for approaches which can learn more robust context-based representations so that the performance on documents containing new unseen key phrases can be improved.

\subsubsection{Domain Adaptation Methods}\label{result:domain_adaptation}
For the lexicon split, one can consider that the training and test data distributions are different, and thus model this as an instance of unsupervised domain adaptation. We evaluate two different approaches for such domain adaptation: adversarial training and multi-task learning.

To address lexical/covariance shift in lexicon dataset split, we adversarially train BiLSTM model with the objective that the resulting document representation will be discriminative for the text classification task while indiscriminative to the training/test domain classification. To perform such domain-invariant adversarial training, we include a discriminator module which consists of three feed-forward layers. To learn the model parameters, we follow the training method of~\newcite{ganin2016domain}, but instead of gradient reversal, we feed the interchanged labels of the domains to fool the discriminator. From the results, we observe that adversarial training of encoder hurts the performance on lexicon datasets.

In multi-task learning, the various tasks which were trained jointly are text classification, denoising autoencoder~\cite{hill2016learning}, and adversarial training. In all these tasks, we share the embedding layer and BiLSTM encoder layer. Denoising autoencoder was trained on the entire dataset using the strategy described in~\newcite{lample2017unsupervised}. From the results, we observe that multi-task learning doesn't help and it further hurts the overall classification accuracy. 

We didn't experiment with these domain adaptation approaches on the random split of the dataset, as it is assumed that it contains the same distribution of lexicons in the training and test set. Also, as the initial results were not promising on lexicon dataset, we didn't try more complex approaches.

\begin{figure}
\centering
\begin{subfigure}{.5\textwidth}
  \centering
  \includegraphics[width=.9\linewidth]{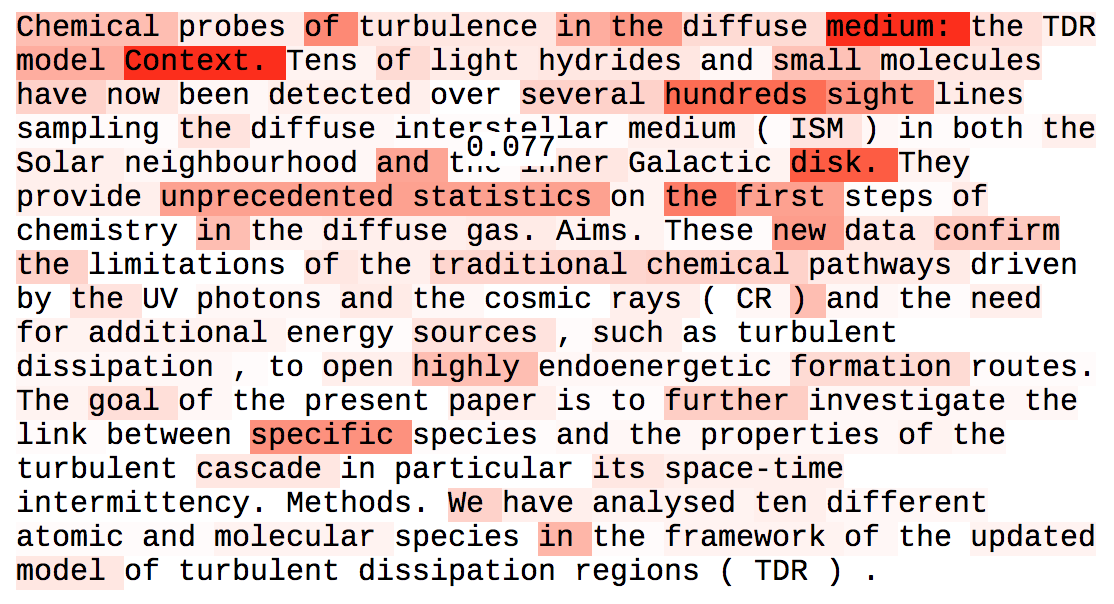}
  \captionsetup{justification=centering}
  \caption{without word anonymization}
  \label{fig:sub1}
\end{subfigure}%
\begin{subfigure}{.5\textwidth}
  \centering
  \includegraphics[width=.9\linewidth]{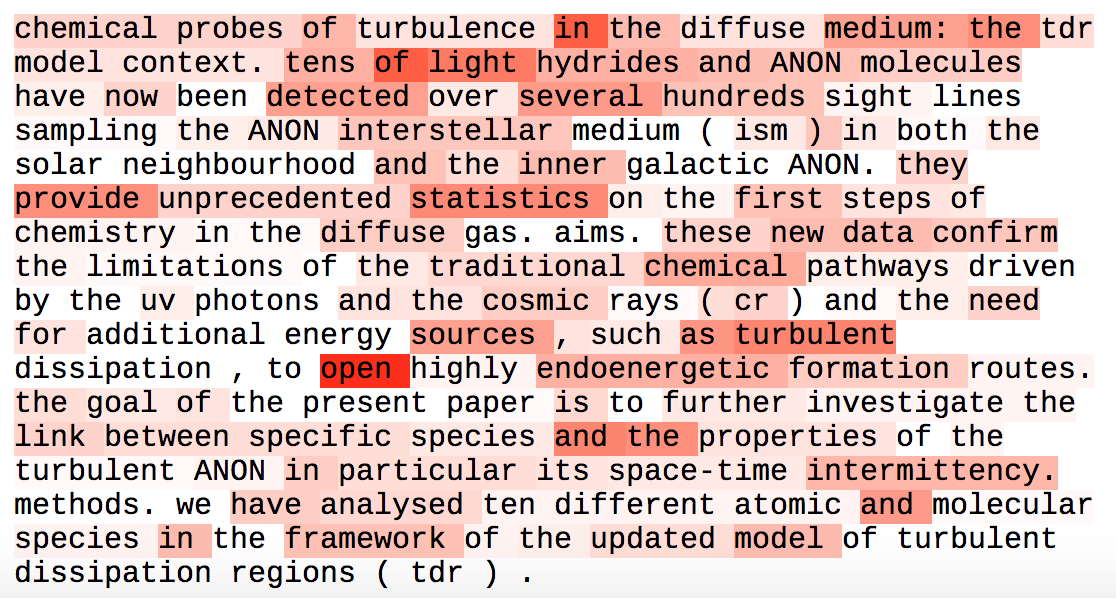}
  \captionsetup{justification=centering}
  \caption{with word anonymization}
  \label{fig:sub2}
\end{subfigure}
  \caption{This attention heat map shows the change in hidden state activations of BiLSTM encoder when some label specific keywords are anonymized for an example instance from Arxiv abstracts dataset whose label is \emph{astro-ph.GA}. In Figure~\ref{fig:sub1}, the trained classifier makes an incorrect prediction as it can be seen that the model is performing a keyword-based classification. In Figure~\ref{fig:sub2} with word anonymization, the attention span also involves other context words and the prediction is correct.}
\label{fig:attn_heatmap}
\end{figure}

\subsection{Our Methods}\label{sec:res_reg}
BiLSTM encoder applied to anonymized training data with random embedding substitution performs around 2\% better on Arxiv abstracts dataset, 2.5\% better on IMDB reviews dataset and gives around 1\% improvement on ACL IMDB lexicon dataset. This shows that keyword anonymization with random embedding substitution can be a good strategy for model regularization in case of lexicon-based split. From a qualitative perspective, we show in Figure~\ref{fig:attn_heatmap} the change of hidden state activations for the BiLSTM encoder after the data is anonymized.

Our approach of adaptive word dropout performs 2.6\% better on Arxiv abstracts, 3.6\% better on IMDB reviews and gives around 2.8\% improvement on ACL IMDB lexicon datasets. We also note that this approach leads to an only marginal drop in accuracy for the random split version of the datasets. One of the reasons why this method is more effective is that word dropout partially masks some lexical terms in the training set thereby lowering the variance of the fitted model.

\section{Conclusion}
Recently, multilayer neural network models have gained wide popularity for text classification tasks due to their much better performance than traditional bag-of-words based approaches. It is widely believed that this happens as neural networks can effectively utilize the word order structure present in documents. But, a potential drawback is that since all neural network approaches are discriminative, they tend to identify key signals in the training data which may not generalize to test data. In this work, we investigate whether these neural network models actually learn to compose the meaning of sentences or just use discriminative keywords.

To test the generalization ability of different state-of-the-art text classifiers, we construct hard datasets where the training and test splits have no direct overlap of lexicons. Our experiments with popular text classifiers show that there is a large drop in test classification accuracy between random and lexicon splits of these datasets. We show that simple regularization techniques such as keyword anonymization can substantially improve the performance of text classifiers. We also observe that adaptive word dropout method which is based on embedding layer's gradient can further improve accuracy and thus reduce the gap between the two dataset splits.

\section*{Acknowledgments}
This work was supported by a generous research funding from CMU, MCDS students grant. We would also like to thank the anonymous reviewers for giving us their valuable feedback which helped to improve the paper.

\bibliographystyle{acl}
\bibliography{coling2018}

\end{document}